\documentclass[runningheads]{llncs}

\usepackage[final,year=2026,ID=xxx]{eccv}

\usepackage{eccvabbrv}

\usepackage{graphicx}
\usepackage{booktabs}
\usepackage{multirow}
\usepackage{subcaption}
\usepackage{bbding} 
\usepackage{amssymb}
\usepackage{pdfpages}

\usepackage[accsupp]{axessibility}  

\usepackage{hyperref}

\usepackage{orcidlink}

\begin{document}

\title{RSGen: Enhancing Layout-Driven Remote Sensing Image Generation with Diverse Edge Guidance} 

\titlerunning{RSGen}


\author{
Xianbao Hou\inst{1,2}\thanks{Equal contribution. \\ 
\textsuperscript{$\dagger$}~Project leader. \\ 
\textsuperscript{\Envelope}~Corresponding authors.} \and
Yonghao He\inst{2}\textsuperscript{$\star, \dagger$} \and
Zeyd Boukhers\inst{3} \and
John See\inst{4} \and
Hu Su\inst{5} \and
Wei Sui\inst{2}\textsuperscript{\Envelope} \and
Cong Yang\inst{1}\textsuperscript{\Envelope}
}

\authorrunning{X. Hou et al.}

\institute{
School of Future Science and Engineering, Soochow University, Suzhou, China \and
D-Robotics, Beijing, China \and
Fraunhofer Institute for Applied Information Technology, Sankt Augustin, Germany
 \and
School of Mathematical and Computing Sciences, Heriot-Watt University Malaysia, Putrajaya, Malaysia
 \and
Institute of Automation, Chinese Academy of Sciences, Beijing, China\\
\email{xbhou2024@stu.suda.edu.cn, wei.sui@d-robotics.cc, cong.yang@suda.edu.cn}
}

\maketitle
\begin{figure}[h]
    \centering    \includegraphics[width=\textwidth]{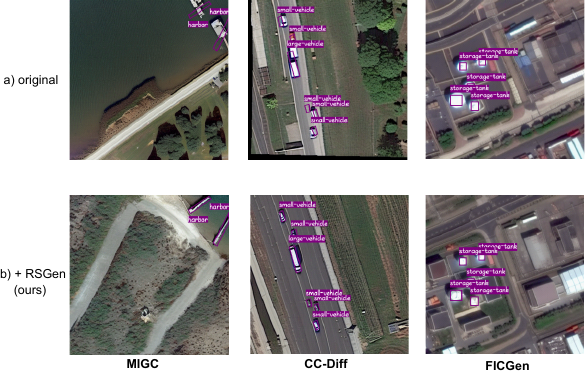}
    \caption{Visualization of controllability between existing L2I methods (MIGC~\cite{zhou2024migc}, CC-Diff~\cite{zhang2024cc}, FICGen~\cite{wang2025ficgen}) and the same methods equipped with RSGen. While the original methods (a) struggle to adhere to the specified bounding boxes, integrating our module (b) significantly enhances instance alignment and control precision.}
    \vspace{-6mm}
    \label{vis_add_ours}
\end{figure}
\begin{abstract}
  Diffusion models have significantly mitigated the impact of annotated data scarcity in remote sensing (RS). 
  Although recent approaches have successfully harnessed these models to enable diverse and controllable Layout-to-Image (L2I) synthesis, they still suffer from limited fine-grained control and fail to strictly adhere to bounding box constraints.
  To address these limitations, we propose RSGen, a plug-and-play framework that leverages diverse edge guidance to enhance layout-driven RS image generation.
  Specifically, RSGen employs a progressive enhancement strategy: 1) it first enriches the diversity of edge maps composited from retrieved training instances via Image-to-Image generation; and 2) subsequently utilizes these diverse edge maps as conditioning for existing L2I models to enforce pixel-level control within bounding boxes, ensuring the generated instances strictly adhere to the layout.
  Extensive experiments across three baseline models demonstrate that RSGen significantly boosts the capabilities of existing L2I models. For instance, with CC-Diff on the DOTA dataset for oriented object detection, we achieve remarkable gains of +9.8/+12.0 in YOLOScore ($\text{mAP}_{50}$/$\text{mAP}_{50-95}$) and +1.6 in mAP on the downstream detection task. Our code will be publicly available: https://github.com/D-Robotics-AI-Lab/RSGen
\end{abstract}

\section{Introduction}
\label{sec:intro}

High-quality remote sensing (RS) data is crucial for improving downstream detection tasks, going beyond specific algorithmic enhancements~\cite{chen2020piou}. As a result, many studies have investigated the potential of augmenting training sets with additional generated data~\cite{tang2025aerogen,ye2026of,zhang2024cc,tang2024crs}. Traditional methods mainly use text prompts to control the semantics of the generated images~\cite{liu2025text2earth,khanna2024diffusionsat}. However, a significant limitation of these approaches is that the generated data often requires manual annotation or additional processing to be effectively used. To mitigate this problem, some studies have focused on enhancing controllability by introducing dense guidance~\cite{tang2024crs,yuan2023efficient}. Unfortunately, these strategies often compromise the diversity of the generated instances.


Recently, layout-to-image (L2I) generation~\cite{zheng2023layoutdiffusion,zhou2024migc,wang2024instancediffusion} has been introduced to achieve a balance between diversity and controllability by using specified object bounding boxes as spatial conditions. While these methods allow for control at the box level and enhance overall diversity, the precision of this control is still limited. As a result, there is often a misalignment between the generated instances and their corresponding bounding boxes~\cref{vis_add_ours} (a). Consequently, even though the generated images may look visually realistic, the corresponding annotations tend to be inaccurate, which reduces their effectiveness for training purposes. In addition to issues with spatial misalignment, current L2I methods do not fully utilize the intrinsic information present in the training samples. They mainly depend on bounding boxes and class labels, neglecting the finer edge details that can contribute to better results.

To address these issues, we introduce RSGen. RSGen consists of two modules: Edge2Edge and L2I FGControl (Frequency-Gated Control with Spatially Gated Injection). The rationale behind RSGen is to incorporate edge maps as auxiliary conditions alongside box-level guidance to achieve pixel-level precision. This ensures that the generated instances align strictly with the specified bounding boxes and improves the quality and reliability of the resulting annotations, as demonstrated in ~\cref{vis_add_ours} (b). Specifically, Edge2Edge retrieves Holistically-Nested Edge Detection (HED)~\cite{xie2015holistically} edge maps conditioned on the input boxes and performs Image-to-Image (I2I) generation with an SDXL~\cite{podell2024sdxl} model fine-tuned via Low-Rank Adaptation (LoRA)~\cite{hu2022lora}, where diversity is promoted by varying seeds and text prompts for downstream data augmentation~\cite{hou2025instada,fan2024divergen}. Furthermore, the L2I FGControl module separates structure from semantics. FGControl extracts high-frequency structural residual features by filtering out low-frequency components, and then injects these residuals only within the bounding boxes via a spatially gated mechanism to keep the guidance spatially confined.

Experiments conducted on three baselines (MIGC~\cite{zhou2024migc}, CC-Diff~\cite{zhang2024cc}, FICGen~\cite{wang2025ficgen}) show consistent improvements in YOLOScore and mean Average Precision (mAP). For example, with CC-Diff on the DIOR-RSVG~\cite{diorrsvg} dataset, we observe enhancements of +3.3/+6.5 in YOLOScore ($\text{mAP}_{50}$/$\text{mAP}_{50-95}$), along with a +0.3 increase in mAP. The improvements are even more significant on the DOTA-v1.0~\cite{xia2018dota} dataset, which focuses on oriented object detection, producing notable increases of +9.8/+12.0 in YOLOScore and +1.6 in mAP.
In summary, our key contributions are as follows:
\begin{itemize}
\item We propose RSGen, a plug-and-play framework that enhances L2I models with fine-grained edge guidance. By providing pixel-level control, RSGen fully harnesses the potential of L2I generation.
\item To tackle the challenge of maintaining generation diversity while utilizing strong pixel-level control, we introduce Edge2Edge. This approach generates diverse edge priors at the source using varied prompts and seeds. Following this, FGControl incorporates these edges as guidance, effectively separating structure from semantics. This allows for precise control through spatial and frequency gating mechanisms.
\item Extensive experiments demonstrate the effectiveness and generalizability of RSGen. It significantly improves spatial alignment precision and enhances downstream detection performance for both horizontal bounding boxes (HBB) and oriented bounding boxes (OBB).
\end{itemize}

\section{Related Work}
This section reviews the literature that forms the basis of our method, focusing on two areas: L2I generation and generative data augmentation.

\subsection{Layout-to-Image Generation}

L2I generation aims to create images based on input bounding boxes and category labels. Early training-free approaches~\cite{jia2024ssmg,xie2023boxdiff,dahary2024yourself} intervene in the attention mechanisms of diffusion models~\cite{rombach2022high,podell2024sdxl} to restrict generation to specific regions, but they suffer from limited controllability. To address these issues, subsequent training-based methods explicitly integrate layout guidance into the model architecture. For example, GLIGEN~\cite{li2023gligen} introduces gated attention to combine spatial information with visual features, achieving strong controllability. In remote sensing, AeroGen~\cite{tang2025aerogen} advances this approach by enabling the generation of oriented bounding boxes, while CC-Diff~\cite{zhang2024cc} enhances coherence between foreground instances and backgrounds. OF-Diff~\cite{ye2026of} utilizes prior masks for improved control, but acquiring these masks can be challenging compared to edge maps, especially for amorphous categories like golf courses. Furthermore, relying solely on simple spatial transformations can limit diversity.


RSGen addresses these limitations in two ways. First, it employs edge guidance that is robust to boundary ambiguity, effectively managing amorphous categories. Second, it utilizes a diffusion model to enhance the diversity of edge priors. By leveraging these diverse edges, our method guides the generation process to achieve both high diversity and precise structural control.

\subsection{Generative Data Augmentation}

Generative data augmentation has become a popular strategy for addressing data scarcity during model training. By leveraging the generative capabilities of diffusion models, this approach generates novel samples to enrich existing datasets, and it is widely adopted across tasks such as classification~\cite{islam2024diffusemix}, object detection~\cite{tang2025aerogen,ye2026of}, and segmentation~\cite{zhao2023x}. Some methods~\cite{fan2024divergen,hou2025instada} employ multi-stage, training-free pipelines, which efficiently avoid the need for model training. However, these methods incur high computational costs due to complex post-processing requirements. In contrast, L2I methods~\cite{zhang2024cc} fine-tune diffusion models to generate instances directly within bounding boxes, which reduces complexity compared to multi-stage processing. This approach facilitates a streamlined augmentation pipeline where generated samples are directly merged into the training dataset. Occasionally, a filtering step is incorporated to ensure quality.


Similarly, our RSGen directly augments the training set, but keeps the base L2I model frozen and only trains the lightweight FGControl module and LoRA~\cite{hu2022lora}, resulting in minimal resource consumption.

\section{Method}
\begin{figure}[t!]
    \centering 
    \includegraphics[width=\textwidth]{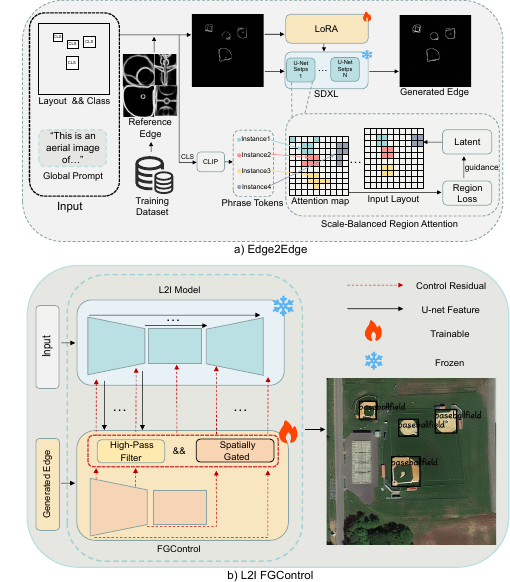}
    \caption{Overview of RSGen, which consists of the Edge2Edge module (a) and the L2I FGControl module (b), where ``CLS'' denotes the class label. The Edge2Edge module enhances the diversity of retrieved edge maps through an I2I process. Subsequently, these diverse edges and layout inputs guide the L2I FGControl module, which interacts with the base L2I model to achieve precise pixel-level control. Our framework significantly increases structural diversity while ensuring fine-grained spatial alignment.}
    \label{fig_overview}
\end{figure}
In this section, we start by reviewing the fundamentals of Latent Diffusion Models (LDMs)~\cite{rombach2022high}. Next, we provide a detailed overview of the RSGen framework (~\cref{fig_overview}), which consists of two key components: the Edge2Edge module, designed for generating diverse edge maps, and the L2I FGControl module, which incorporates edge guidance to ensure accurate layout alignment. Together, these components address the challenges of limited diversity and spatial misalignment in remote sensing image generation.
\begin{figure}[t!]
    \centering    \includegraphics[width=\textwidth]{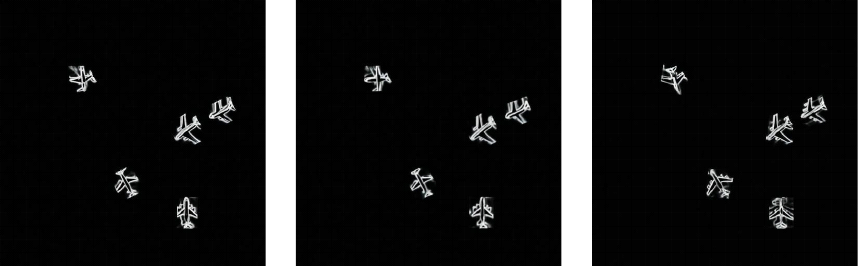}
    \caption{Visualization of diverse edge maps generated by the Edge2Edge module. Our method employs distinct random seeds to generate varied structural details within the specified bounding boxes, significantly enhancing the diversity of the structural priors.}
    \label{fig_diverse_edge}
\end{figure}
\subsection{Preliminary}
Denoising Diffusion Probabilistic Models~\cite{ho2020denoising} synthesize images by iteratively removing Gaussian noise directly in pixel space. In contrast, LDMs~\cite{rombach2022high} significantly accelerate this process within a compressed latent space constructed by a Variational Autoencoder (VAE)~\cite{kingma2013auto}. Specifically, an input image $x$ is first encoded into a latent representation $z_0 = \mathcal{E}(x)$ by a pre-trained VAE encoder $\mathcal{E}$. Gaussian noise is then injected into $z_0$ to yield a noisy state $z_t = \alpha_t z_0 + \sigma_t \epsilon$, with $\epsilon \sim \mathcal{N}(0,I)$, where $\alpha_t$ and $\sigma_t$ are coefficients determined by the noise schedule.
A denoising U-Net~\cite{ronneberger2015u} $\epsilon_\theta$ is optimized to estimate the added noise $\epsilon$ conditioned on the timestep $t$ ($t\in\{1,\ldots,T\}$) and auxiliary context $c$ (e.g., layouts and text). 
The LDM loss, denoted as $\mathcal{L}_{LDM}$, is defined to minimize the Mean Squared Error (MSE) between the predicted noise and the actual noise, which can be expressed as follows:
\begin{equation}
    \mathcal{L}_{LDM} = \mathbb{E}_{z_0,\epsilon \sim \mathcal{N}(0, I),t, c} \left[ \| \epsilon - \epsilon_\theta(z_t, t, c) \|_2^2 \right].
\end{equation}

\subsection{Edge2Edge}
As highlighted in~\cref{fig_overview} (a), we construct a database of reference edges using HED~\cite{xie2015holistically} and retrieve the optimal candidates based on the class and aspect ratio of each input bounding box. 
The retrieved edges are then assembled into a composite edge map, which serves as the input for a fine-tuned SDXL~\cite{podell2024sdxl} model within an I2I process.
To ensure that the generated edge map aligns with the specified bounding boxes, we incorporate a novel Scale-Balanced Region Attention mechanism. Additionally, by utilizing prompts derived from input instance classes and various random seeds, this module generates a diverse edge map, as illustrated in ~\cref{fig_diverse_edge}.


\subsubsection{\textbf{LoRA Fine-tuning and Scale-Balanced Region Attention.}}
The base SDXL model is fine-tuned via LoRA~\cite{hu2022lora} to align with the visual domain of HED edges. 
By utilizing composite maps assembled from retrieved references as input, the model parameters are optimized to shift the distribution toward the target HED style. 
This parameter-efficient strategy ensures that the generated edges maintain high fidelity to the desired structural patterns without the computational cost of full-parameter training.

However, although the fine-tuned SDXL model captures the visual characteristics of HED edges, standard I2I generation still lacks explicit spatial constraints. 
This limitation leads to two critical issues: 1) Semantic Misalignment, where a single global prompt fails to restrict semantic attributes to their corresponding bounding boxes; and 2) Boundary Overflow, where generated structural edges drift beyond the limits of the input layout boxes. 
Such unconstrained generation produces noisy priors that significantly degrade the performance of the subsequent L2I FGControl module.

Inspired by training-free layout control methods~\cite{dahary2024yourself,xie2023boxdiff}, we propose Scale-Balanced Region Attention that intervenes in the U-Net~\cite{ronneberger2015u} denoising process to enforce area-aware constraints, effectively mitigating the optimization bias toward larger bounding boxes.

\begin{figure}[t!]
    \centering
    \includegraphics[width=\textwidth]{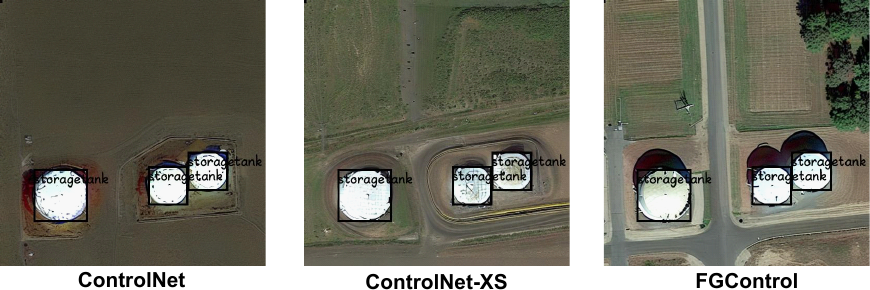}
    \caption{Comparison of ControlNet~\cite{zhang2023adding}, ControlNet-XS~\cite{zavadski2024controlnet}, and our FGControl. Standard global control methods suffer from feature entanglement, causing background chaos. Conversely, FGControl strictly confines high-frequency structural guidance within the layout bounding boxes, achieving fine-grained local control without interfering with the global semantic synthesis.}
    \label{fig_compare_control}
\end{figure}

\subsubsection{(i) Latent Guidance via Region Loss.}
In the early timesteps of the denoising process, we perform iterative latent updates to align the semantic concepts with their target spatial regions.
To achieve this, we construct a comprehensive text prompt by concatenating the class names of all layout instances into the format: ``\textit{hed edge map. [class 1], [class 2]...}''.
Subsequently, using the CLIP~\cite{radford2021learning} text encoder, we tokenize the prompt and identify the specific token indices $K_i$ corresponding to the class name of each bounding box $b_i$.

During the U-Net forward pass, we extract both cross-attention and self-attention maps from designated low-resolution layers where layout features are concentrated. For cross-attention, let $K_i$ denote the set of token indices corresponding to the class name of instance $i$. We obtain the aggregated spatial attention map $\mathcal{A}_{i} \in \mathbb{R}^{H \times W}$ ($H, W$ are latent dimensions) by summing over all tokens $k \in K_i$ and averaging across all attention heads. For self-attention, $\mathcal{A}_{i}$ is directly derived from the spatial region interactions without text tokens.

For regional alignment, let $\mathcal{M}^{fg}_i \in \{0, 1\}^{H \times W}$ be the binary mask of bounding box $b_i$, and $\mathcal{M}^{bg} \in \{0, 1\}^{H \times W}$ be the global background mask. We compute the mean foreground activation ($A^{fg}_i$) and background leakage ($A^{bg}_i$):

\begin{equation}
    A^{fg}_i = \frac{\sum_{x,y} \left( \mathcal{A}_{i}^{(x,y)} \odot \mathcal{M}^{fg, (x,y)}_i \right)}{\sum_{x,y} \mathcal{M}^{fg, (x,y)}_i + \epsilon}, \quad 
    A^{bg}_i = \frac{\sum_{x,y} \left( \mathcal{A}_{i}^{(x,y)} \odot \mathcal{M}^{bg, (x,y)} \right)}{\sum_{x,y} \mathcal{M}^{bg, (x,y)} + \epsilon}.
\end{equation}
where $\odot$ denotes element-wise multiplication, $(x,y)$ represents the spatial pixel coordinates, and $\epsilon = 10^{-6}$ is a small constant.

The region loss $\mathcal{L}_{reg}$ is an Intersection over Union (IoU)-inspired objective to maximize foreground response and penalize background leakage. Computed per step and averaged across targeted layers, it is defined as:
\begin{equation}
    \mathcal{L}_{reg} = \sum_{i=1}^{N} \left( 1 - \frac{A^{fg}_i}{A^{fg}_i + N \cdot A^{bg}_i} \right)^2 .
\end{equation}
where $N$ is the total number of bounding boxes. Crucially, scaling background leakage by $N$ counters the area imbalance between instances and the global background, preventing it from dominating optimization.

With this loss, we compute its gradient with respect to the input noisy latent $z_t$ and perform a gradient descent update: $z_t \leftarrow z_t - \lambda \nabla_{z_t} \mathcal{L}_{reg}$, where $\lambda$ is a step size dynamically decreasing from $8$ to $2$.
Crucially, this updated latent dynamically shapes the subsequent attention maps. 
Since the spatial Query features ($Q$) in the cross-attention layers are directly projected from the latent $z_t$, shifting $z_t$ along the negative gradient geometrically translates the high-activation visual features. 
When the modified $z_t$ is fed back into the U-Net, the resulting attention maps naturally concentrate the semantic response within the bounding boxes.
\subsubsection{(ii) Region-Masked Attention.}
While the early timesteps establish the global layout via latent updates, the later steps require strict spatial confinement. To prevent boundary overflow, the attention computation is modified by introducing a masking matrix $\mathbf{M}$ to both cross-attention and self-attention layers.
Based on the input layout, $\mathbf{M}_{ij}$ is set to $0$ if the query-key pair $(i, j)$ aligns with the same instance, and $-\infty$ otherwise.
The attention score is then reformulated:
\begin{equation}
    \text{Attention}(Q, K, V) = \text{Softmax}\left( \frac{Q K^T}{\sqrt{d}} + \mathbf{M} \right) V.
\end{equation}
By adding this large negative mask before the Softmax operation, information flow between unrelated regions is physically blocked. The design intentionally forces the model to focus on adhering to the strict edge constraints within the specified bounding boxes.

To translate the diverse edge priors into precise pixel-level control, we introduce the L2I FGControl module (~\cref{fig_overview} (b)). 
While retaining the lightweight efficiency of ControlNet-XS~\cite{zavadski2024controlnet}, FGControl fundamentally addresses the issues of standard global injections, which often cause background chaos and overwhelming structural control in L2I tasks, as shown in \cref{fig_compare_control}. 
To explicitly decouple structure from semantics, FGControl incorporates the generated edge maps as auxiliary conditions and utilizes a high-pass filter to extract high-frequency structural residuals. 
Through a spatially gated mechanism, these residuals are explicitly injected into the base model. 
This synergistic interaction ensures that the pixel-level structural guidance is strictly confined within the layout bounding boxes without affecting the background.
\begin{table}[t]
\caption{Quantitative comparison of baseline models with and without our proposed RSGen on DIOR-RSVG and DOTA datasets. Models equipped with RSGen achieve substantial improvements in layout consistency (YOLOScore) alongside highly competitive generation fidelity (FID).}
\centering
\setlength{\tabcolsep}{8pt} 
\begin{tabular}{l ccc c ccc}
\toprule
\multirow{3}{*}{Method} & \multicolumn{3}{c}{DIOR-RSVG (HBB)} & \phantom{a} & \multicolumn{3}{c}{DOTA (OBB)} \\
\cmidrule{2-4} \cmidrule{6-8}
& \multirow{2}{*}{FID $\downarrow$} & \multicolumn{2}{c}{YOLOScore $\uparrow$} & & \multirow{2}{*}{FID $\downarrow$} & \multicolumn{2}{c}{YOLOScore $\uparrow$} \\
\cmidrule{3-4} \cmidrule{7-8}
& & $\text{mAP}_{50}$ & $\text{mAP}_{50-95}$ & & & $\text{mAP}_{50}$ & $\text{mAP}_{50-95}$ \\
\midrule
MIGC~\cite{zhou2024migc}         & \textbf{79.32} & 63.2 & 38.4 & & \textbf{66.80} & 53.0 & 30.0 \\
+ Ours                     & 85.04 & \textbf{68.7} & \textbf{45.8} & & 68.07 & \textbf{55.1} & \textbf{37.7} \\
\midrule
CC-Diff~\cite{zhang2024cc}       & \textbf{66.75} & 66.8 & 41.2 & & 47.72 & 57.7 & 34.0 \\
+ Ours                     & 68.12 & \textbf{70.1} & \textbf{47.7} & & \textbf{46.86} & \textbf{67.5} & \textbf{46.0} \\
\midrule
FICGen~\cite{wang2025ficgen}     & 74.26 & 64.9 & 39.7 & & \textbf{48.46} & 74.0 & 49.3 \\
+ Ours                     & \textbf{73.26} & \textbf{65.7} & \textbf{42.2} & & 48.56 & \textbf{75.2} & \textbf{53.2} \\
\bottomrule
\label{table:compare_ours}
\end{tabular}
\end{table}
\subsection{L2I FGControl}
\subsubsection{(i) Frequency Gated Structure Decoupling.} 
The fundamental goal of FGControl is to provide precise geometric edge guidance without interfering with the semantic synthesis of the base model. Since structural edges naturally correspond to high-frequency features, we propose a frequency-aware decoupling strategy to explicitly extract sharp edge features. Given a control residual $h_{res}$ from the control branch, we define the intermediate residual map as $\Delta h = \text{ZeroConv}(h_{res})$. We then apply a Fast Fourier Transform ($\mathcal{F}$) and a high-pass filter $\mathcal{H}$ to extract high-frequency structural edges:
\begin{equation} 
    \Delta h_{high} = \mathcal{F}^{-1} \left( \mathcal{F}(\Delta h) \odot \mathcal{H} \right). 
\end{equation} 
Specifically, $\mathcal{H}$ zeroes out a central $\frac{H}{d} \times \frac{W}{d}$ low-frequency band ($d=16$). We then apply soft thresholding ($\tau = 0.05$) to eliminate negligible noise, yielding the purified structural residual $\Delta h_{str}$:
\begin{equation} 
    \Delta h_{str} = \operatorname{sgn}(\Delta h_{high}) \odot \max(|\Delta h_{high}| - \tau, 0). 
\end{equation} 
By completely discarding the low-frequency features, FGControl strictly dictates structural layouts while the base model governs the semantic content.

\subsubsection{(ii) Spatially Gated Injection.} 
Beyond frequency decoupling, strict spatial confinement is imperative to prevent structural artifacts from leaking into the background. 
For each instance, we dynamically resize its bounding box mask $\mathcal{M}^{fg}$ to match the spatial resolution of the current U-Net~\cite{ronneberger2015u} layer. 
A spatially gated mechanism is then formulated by explicitly multiplying the purified high-frequency residuals with this mask. The final injected residual is formulated as:
\begin{equation}
    \Delta h_{final} = \Delta h_{str} \odot \mathcal{M}^{fg}.
\end{equation}
Consequently, the base L2I model receives these precise structural constraints exclusively within the layout regions, achieving fine-grained pixel-level alignment while leaving the background generation completely unaffected.

\section{Experiments}
In this section, we conduct comprehensive experiments to assess the effectiveness and generalization of RSGen. We incorporate our framework into various L2I baselines and evaluate its performance across a range of remote sensing datasets, focusing on both the quality of generation and its application in downstream object detection tasks. Additional implementation details, efficiency analysis, supplementary experiments, and limitations are provided in the Appendix.
\subsection{Experimental Settings}
\noindent{\textbf{Datasets.}}
Following  CC-Diff~\cite{zhang2024cc}, we evaluate our proposed RSGen on two widely used remote sensing datasets: DIOR-RSVG~\cite{diorrsvg} and DOTA-v1.0~\cite{xia2018dota}. 
Constructed based on the large-scale DIOR~\cite{li2020object} dataset, DIOR-RSVG provides HBB annotations exclusively, serving as our primary benchmark for horizontal object generation. 
In contrast, DOTA-v1.0 is a challenging dataset comprising 15 categories, featuring dense scenes and small objects. We crop the images from DOTA to 512 × 512 following CC-Diff.
Furthermore, we utilize this dataset to conduct a comprehensive dual verification for both HBB and OBB detection.

\noindent{\textbf{Benchmarks.}}
We evaluate our method across three L2I generation models originally designed for distinct domains: MIGC~\cite{zhou2024migc} for natural images, CC-Diff for remote sensing, and FICGen~\cite{wang2025ficgen} for degraded scenes.

\begin{table}[t]
\caption{Comparison of downstream object detection performance. By mixing synthetic and real data at a 1:1 ratio, models trained with data generated by RSGen achieve overall accuracy improvements across both HBB and OBB settings.}
\label{table:trainability}
\centering
\setlength{\tabcolsep}{5pt} 
\resizebox{\linewidth}{!}{
    \begin{tabular}{l ccc c ccc c cc}
    \toprule
    \multirow{3}{*}{Method} & \multicolumn{3}{c}{DIOR-RSVG (HBB)} & \phantom{a} & \multicolumn{3}{c}{DOTA (HBB)} & \phantom{a} & \multicolumn{2}{c}{DOTA (OBB)} \\
    \cmidrule{2-4} \cmidrule{6-8} \cmidrule{10-11}
    & mAP & $\text{mAP}_{50}$ & $\text{mAP}_{75}$ & & mAP & $\text{mAP}_{50}$ & $\text{mAP}_{75}$ & & $\text{mAP}_{50}$ & $\text{mAP}_{75}$ \\
    \midrule
    MIGC~\cite{zhou2024migc}         & 54.3 & \textbf{78.9} & 59.8 & & 38.3 & \textbf{64.5} & 39.2 & & \textbf{53.55} & 19.68 \\
    + Ours                     & \textbf{55.0} & 78.7 & \textbf{61.1} & & \textbf{38.9} & \textbf{64.5} & \textbf{40.5} & & 53.19 & \textbf{21.92} \\
    \midrule
    CC-Diff~\cite{zhang2024cc}       & 54.7 & 78.4 & 60.1 & & 37.4 & 63.2 & 38.6 & & 54.80 & 20.11 \\
    + Ours                     & \textbf{55.0} & \textbf{78.8} & \textbf{60.7} & & \textbf{39.0} & \textbf{64.2} & \textbf{40.4} & & \textbf{54.92} & \textbf{21.65} \\
    \midrule
    FICGen~\cite{wang2025ficgen}     & 54.5 & 78.7 & 60.7 & & 39.2 & \textbf{64.8} & 40.8 & & 55.81 & 22.66 \\
    + Ours                     & \textbf{54.6} & \textbf{78.8} & \textbf{61.1} & & \textbf{39.5} & \textbf{64.8} & \textbf{41.4} & & \textbf{55.93} & \textbf{23.03} \\
    \bottomrule
    \end{tabular}
}
\end{table}

\noindent{\textbf{Implementation.}}
We train CC-Diff and FICGen following their official settings, while MIGC is fine-tuned using its official weights under CC-Diff's settings, except for a reduced 50 training epochs. 
For FGControl, we freeze the base L2I models. Images are resized to 512 × 512, and the module is optimized for 75,000 steps with a batch size of 8 and a fixed learning rate of 8e-5.

\noindent{\textbf{Metrics.}}
To comprehensively evaluate the performance of RSGen, we assess the generated images from three distinct perspectives: \textbf{(1) Fidelity.} We utilize the Fréchet Inception Distance (FID)~\cite{heusel2017gans} to assess the perceptual quality of the generated images. FID measures the distributional distance between generated and real features, providing a robust evaluation of contextual coherence. \textbf{(2) Layout Consistency.} The YOLOScore~\cite{li2021image} is employed to quantify the alignment between generated instances and spatial layouts. Specifically, we fine-tune a YOLOv8 detector~\cite{yolov8} for HBB and a YOLOv8-OBB detector for OBB. The precision on the generated images directly reflects the control capability. \textbf{(3) Trainability.} To evaluate the effect of RSGen for data augmentation, we mix synthetic images with real training data to train downstream detectors. We employ Faster R-CNN~\cite{ren2016faster} for HBB tasks (reporting mAP, mAP$_{50}$, mAP$_{75}$.) and Oriented Faster R-CNN~\cite{xie2021oriented} for OBB tasks (reporting mAP$_{50}$, mAP$_{75}$).

\begin{table}[t]
\caption{Comparison of downstream object detection performance under different real-to-synthetic data ratios. All experiments are conducted on the DOTA dataset utilizing the CC-Diff~\cite{zhang2024cc} baseline.}
\label{table:ratios}
\centering
\setlength{\tabcolsep}{8pt} 
\resizebox{\linewidth}{!}{
    \begin{tabular}{lc ccc c cc}
    \toprule
    \multirow{2}{*}{Method} & \multirow{2}{*}{Ratio (Real:Syn)} & \multicolumn{3}{c}{DOTA (HBB)} & \phantom{a} & \multicolumn{2}{c}{DOTA (OBB)} \\
    \cmidrule{3-5} \cmidrule{7-8}
    & & mAP & $\text{mAP}_{50}$ & $\text{mAP}_{75}$ & & $\text{mAP}_{50}$ & $\text{mAP}_{75}$ \\
    \midrule
    CC-Diff~\cite{zhang2024cc}  & \multirow{2}{*}{1:3} & \textbf{39.0} & \textbf{65.2} & 39.8 & & 55.72 & 21.11 \\
    \quad + Ours                &                      & \textbf{39.0} & 64.2 & \textbf{40.0} & & \textbf{56.07} & \textbf{24.36} \\
    \midrule
    CC-Diff~\cite{zhang2024cc}  & \multirow{2}{*}{1:2} & 38.8 & \textbf{65.4} & 39.7 & & 54.75 & 21.44 \\
    \quad + Ours                &                      & \textbf{39.3} & 64.4 & \textbf{41.2} & & \textbf{55.98} & \textbf{22.34} \\
    \midrule
    CC-Diff~\cite{zhang2024cc}  & \multirow{2}{*}{1:1} & 37.4 & 63.2 & 38.6 & & 54.80 & 20.11 \\
    \quad + Ours                &                      & \textbf{39.0} & \textbf{64.2} & \textbf{40.4} & & \textbf{54.92} & \textbf{21.65} \\
    \midrule
    CC-Diff~\cite{zhang2024cc}  & \multirow{2}{*}{0:1} & 16.2 & 30.9 & 14.7 & & 22.07 & 9.49 \\
    \quad + Ours                &                      & \textbf{21.9} & \textbf{38.6} & \textbf{21.3} & & \textbf{33.24} & \textbf{12.16} \\
    \bottomrule
    \end{tabular}
}
\end{table}
\subsection{Main Results}
\textbf{Observation 1: RSGen significantly enhances layout consistency with only marginal fluctuations in generation fidelity.} As shown in~\cref{table:compare_ours}, we evaluate the effectiveness by integrating RSGen into three baselines (MIGC, CC-Diff, and FICGen). While the enforcement of strict structural edge constraints leads to slight increases in FID in certain cases (e.g., MIGC on DIOR-RSVG), the overall quality remains highly competitive, and in some instances (e.g., CC-Diff on DOTA), the FID even improves. Specifically, whether evaluated under the HBB setting on DIOR-RSVG or the OBB setting on DOTA, the models equipped with our module achieve substantial improvements in YOLOScore compared to their original counterparts. Notably, the improvements in the comprehensive $\text{mAP}_{50-95}$ metric are even more pronounced than those in $\text{mAP}_{50}$. For instance, integrating RSGen into CC-Diff yields remarkable $\text{mAP}_{50-95}$ gains of +6.5 on DIOR-RSVG and +12.0 on DOTA. Such substantial increases strongly indicate that our method significantly enhances the fine-grained control precision within the bounding boxes. This consistent performance boost across distinct baselines confirms that our module effectively trades negligible variances in global image distribution for precise, pixel-level layout alignment.

\noindent\textbf{Observation 2: RSGen significantly boosts the accuracy of downstream object detection tasks.} Following the experimental settings of CC-Diff, we mix the synthetic images generated by our method with the real training data at a 1:1 ratio. As shown in~\cref{table:trainability}, integrating RSGen into the base models leads to broad increases in the standard mAP under the HBB setting on both the DIOR-RSVG and DOTA datasets. Notably, when comparing different IoU thresholds, the improvements in $\text{mAP}_{75}$ are consistently more pronounced than those in $\text{mAP}_{50}$. For example, adding RSGen to CC-Diff on the DOTA dataset yields a substantial +1.8 gain in $\text{mAP}_{75}$, compared to a +1.0 gain in $\text{mAP}_{50}$.

To provide a deeper comparison, we further evaluate the models under the more challenging OBB setting on DOTA. In this stringent scenario, the gains in $\text{mAP}_{50}$ are relatively marginal; for instance, CC-Diff equipped with our module only shows a slight +0.12 increase over the baseline. However, the performance boost primarily manifests in the stricter $\text{mAP}_{75}$ metric, where it achieves a notable +1.54 improvement. These outsized gains at higher IoU thresholds across both HBB and OBB tasks conclusively demonstrate that our method provides substantially stronger and more precise pixel-level control capabilities.

\begin{figure}[t]
    \centering    \includegraphics[width=\textwidth]{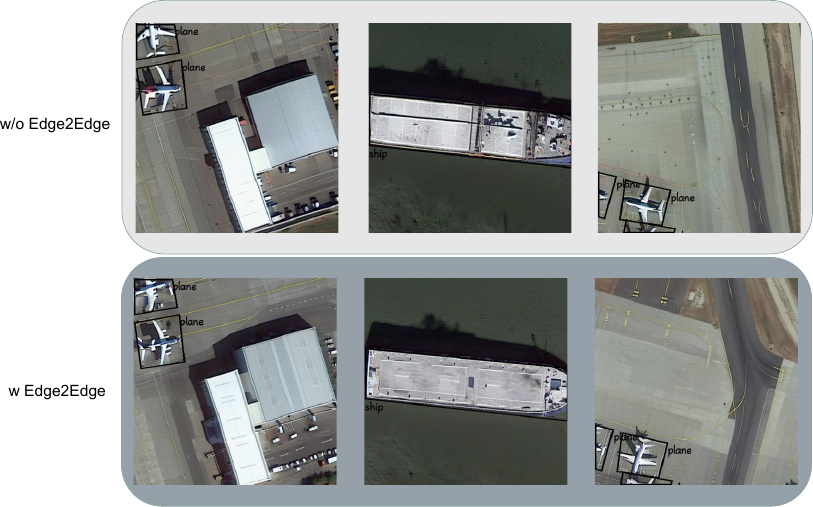}
    \caption{Qualitative comparison of generated instances with and without the Edge2Edge module. Incorporating the Edge2Edge module introduces rich structural variations, significantly enhancing both the structural and overall diversity of the generated instances within the specified bounding boxes.}
    \label{fig_ablation_edge2edge}
\end{figure}

\noindent\textbf{Observation 3: RSGen consistently improves downstream detection performance across various synthetic data ratios.} 
To further validate the robustness and scalability of our method, we conduct extended experiments using the CC-Diff baseline on the DOTA dataset. Specifically, we train the detectors under both HBB and OBB settings using different synthetic-to-real data mixing ratios, including 1:3, 1:2, and purely synthetic data. 
As reported in~\cref{table:ratios}, the evaluation across different data mixing proportions reveals several insightful trends. 
First, under the HBB setting, simply increasing the proportion of synthetic data from 1:2 to 1:3 does not yield continuous performance gains. 
However, our method at a minimal 1:1 ratio ($\text{mAP}$ 39.0, $\text{mAP}_{75}$ 40.4) matches or exceeds the performance of the baseline trained with a heavily augmented 1:3 ratio ($\text{mAP}$ 39.0, $\text{mAP}_{75}$ 39.8). This comparison highlights the exceptional data efficiency of our method, demonstrating that high-quality, strictly aligned instances can saturate the HBB detection performance with much less data volume.

Conversely, under the OBB setting, increasing the synthetic data ratio yields sustained improvements. 
Compared to the baseline, RSGen boosts $\text{mAP}_{50}$ (e.g., +1.23 at 1:2 ratio), while the gains are even more pronounced in the stricter $\text{mAP}_{75}$, peaking at 24.36 (+3.25) under the 1:3 ratio. 
This shows that for complex oriented object detection tasks, our fine-grained control ensures strict spatial alignment, providing highly accurate annotations for detector training.
In the extreme purely synthetic setting, incorporating our module leads to substantial performance gains over the baseline. 
Specifically, the detector achieves massive increases of +5.7 in mAP and +6.6 in $\text{mAP}_{75}$ for HBB, alongside an impressive +11.17 surge in $\text{mAP}_{50}$ and +2.67 in $\text{mAP}_{75}$ for OBB. 
Ultimately, these results convincingly confirm that RSGen generates highly accurate and structurally reliable instances, maximizing both training efficiency and fine-grained spatial alignment for downstream applications.

\begin{table}[t]
\caption{Ablation of the Edge2Edge module on DIOR-RSVG. While maintaining nearly identical generation fidelity (FID), the module significantly boosts layout consistency (YOLOScore), particularly in the stricter IoU threshold.}
\label{table:ablation_edge_indomain}
\centering
\setlength{\tabcolsep}{24pt}
\begin{tabular}{l ccc}
\toprule
\multirow{2}{*}{Method} & \multirow{2}{*}{FID $\downarrow$} & \multicolumn{2}{c}{YOLOScore $\uparrow$} \\
\cmidrule{3-4}
& & $\text{mAP}_{50}$ & $\text{mAP}_{50-95}$ \\
\midrule
w/o Edge2Edge & 68.40 & 69.5 & 46.9 \\
w/ Edge2Edge  & \textbf{68.12} & \textbf{70.1} & \textbf{47.7} \\
\bottomrule
\end{tabular}
\end{table}

\subsection{Ablation Study}

\noindent\textbf{Effect of the Edge2Edge Module.} Edge2Edge enriches structural diversity, as qualitatively shown in~\cref{fig_ablation_edge2edge}. As reported in~\cref{table:ablation_edge_indomain}, it significantly improves the YOLOScore, particularly in the comprehensive $\text{mAP}_{50-95}$ metric (+0.8). Crucially, the FID remains nearly identical (68.12 and 68.40), demonstrating that diverse structural priors effectively enhance instance-layout alignment and realism without compromising generation fidelity of the base model.

To verify that the structural variations introduce meaningful diversity rather than noise, we conduct a cross-dataset generalization test by training on DIOR-RSVG and evaluating on DOTA. As shown in~\cref{table:cross_dataset}, under this challenging setting, the model equipped with Edge2Edge consistently outperforms the baseline across most categories. These results confirm that the structural variations introduced by Edge2Edge enhance cross-dataset generalization, highlighting its robustness across domains and its importance for practical applications.

\noindent\textbf{Effect of the FGControl Module.} To validate the design of the FGControl module, we progressively ablate its key components.
As shown in~\cref{table:ablation_fgcontrol}, our baseline model is ControlNet-XS~\cite{zavadski2024controlnet}, a typical global control mechanism.
However, this global injection forces structural conditions into background regions, causing severe background confusion and degraded image quality.

By introducing the Spatially Gated mechanism, we confine the residual features within the target bounding boxes, which effectively eliminates background noise and improves generation fidelity. Finally, the high-pass filter decouples structural guidance from semantic features. Despite a slight FID increase due to intensified structural focus, the complete FGControl module achieves the highest layout consistency ($\text{mAP}_{50-95}$ 46.9), demonstrating its distinct advantage in enabling precise, fine-grained control while preserving background coherence.

\begin{table}[t]
\caption{Cross-dataset validation. Models are trained on DIOR-RSVG and evaluated on DOTA. The Edge2Edge module improves AP across most categories, demonstrating its strong generalization  capability, which is crucial for practical applications.}
\label{table:cross_dataset}
\centering
\setlength{\tabcolsep}{2pt}
\resizebox{\linewidth}{!}{
    \begin{tabular}{l c ccccccc} 
    \toprule
    Method  & vehicle & ship & basketballcourt & groundtrackfield & harbor & tenniscourt & airplane \\
    \midrule
    w/o Edge2Edge  & 6.8 & \textbf{8.3} & 14.9 & \textbf{13.6} & 9.6 & 42.6 & 8.2 \\
    w/ Edge2Edge   & \textbf{7.1} & \textbf{8.3} & \textbf{15.5} & 13.5 & \textbf{10.2} & \textbf{44.6} & \textbf{8.4} \\
    \bottomrule
    \end{tabular}
}
\end{table}

\begin{table}[t]
\caption{Ablation of the FGControl module on the DIOR-RSVG dataset. The spatially gated mechanism resolves background confusion, while the high-pass filter further decouples features to achieve the highest layout consistency.}
\label{table:ablation_fgcontrol}
\centering
\setlength{\tabcolsep}{16pt} 
\resizebox{\linewidth}{!}{ 
    \begin{tabular}{l c cc}
    \toprule
    \multirow{2}{*}{Method} & \multirow{2}{*}{FID $\downarrow$} & \multicolumn{2}{c}{YOLOScore $\uparrow$} \\
    \cmidrule{3-4}
    & & $\text{mAP}_{50}$ & $\text{mAP}_{50-95}$ \\
    \midrule
    Baseline (Global Control) & 112.43 & 62.1 & 45.5 \\
    + Spatially Gated         & \textbf{64.38} & 68.7 & 45.7 \\
    + High-pass Filter (Full) & 68.40 & \textbf{69.5} & \textbf{46.9} \\
    \bottomrule
    \end{tabular}
}
\end{table}
\section{Conclusion}
In this paper, we introduced RSGen, a novel plug-and-play framework designed to resolve the misalignment between generated instances and given bounding boxes in L2I generation. Within this framework, the Edge2Edge module enriches the structural diversity of the instances and significantly boosts the generalization capability of the model. Building upon this, the L2I FGControl module leverages these diverse edge priors to achieve precise, pixel-level layout control. 

\par\vfill\par


%
%
\bibliographystyle{splncs04}
\bibliography{main}

@String(AAAI  = {AAAI})

@inproceedings{chen2020piou,
  title={Piou loss: Towards accurate oriented object detection in complex environments},
  author={Chen, Zhiming and Chen, Kean and Lin, Weiyao and See, John and Yu, Hui and Ke, Yan and Yang, Cong},
  booktitle={European conference on computer vision},
  pages={195--211},
  year={2020},
  organization={Springer}
}

@article{zhang2024cc,
  title={CC-Diff: enhancing contextual coherence in remote sensing image synthesis},
  author={Zhang, Mu and Liu, Yunfan and Liu, Yue and Zhao, Yuzhong and Ye, Qixiang},
  journal={arXiv preprint arXiv:2412.08464},
  year={2024}
}

@inproceedings{tang2025aerogen,
  title={AeroGen: Enhancing remote sensing object detection with diffusion-driven data generation},
  author={Tang, Datao and Cao, Xiangyong and Wu, Xuan and Li, Jialin and Yao, Jing and Bai, Xueru and Jiang, Dongsheng and Li, Yin and Meng, Deyu},
  booktitle={Proceedings of the Computer Vision and Pattern Recognition Conference},
  pages={3614--3624},
  year={2025}
}

@inproceedings{ye2026of,
title={Object Fidelity Diffusion for Remote Sensing Image Generation},
author={Ye, Ziqi and Ma, Shuran and Yang, Jie and Yang, Xiaoyi and Gong, Ziyang and Yang, Xue and Wang, Haipeng},
booktitle={The Fourteenth International Conference on Learning Representations},
year={2026},
url={https://openreview.net/forum?id=ngfIm9aPsH}
}

@article{tang2024crs,
  title={Crs-diff: Controllable remote sensing image generation with diffusion model},
  author={Tang, Datao and Cao, Xiangyong and Hou, Xingsong and Jiang, Zhongyuan and Liu, Junmin and Meng, Deyu},
  journal={IEEE Transactions on Geoscience and Remote Sensing},
  year={2024},
  publisher={IEEE}
}

@inproceedings{
khanna2024diffusionsat,
title={DiffusionSat: A Generative Foundation Model for Satellite Imagery},
author={Samar Khanna and Patrick Liu and Linqi Zhou and Chenlin Meng and Robin Rombach and Marshall Burke and David B. Lobell and Stefano Ermon},
booktitle={The Twelfth International Conference on Learning Representations},
year={2024},
url={https://openreview.net/forum?id=I5webNFDgQ}
}

@article{liu2025text2earth,
  title={Text2earth: Unlocking text-driven remote sensing image generation with a global-scale dataset and a foundation model},
  author={Liu, Chenyang and Chen, Keyan and Zhao, Rui and Zou, Zhengxia and Shi, Zhenwei},
  journal={IEEE Geoscience and Remote Sensing Magazine},
  year={2025},
  publisher={IEEE}
}

@article{yuan2023efficient,
  title={Efficient and controllable remote sensing fake sample generation based on diffusion model},
  author={Yuan, Zhiqiang and Hao, Chongyang and Zhou, Ruixue and Chen, Jialiang and Yu, Miao and Zhang, Wenkai and Wang, Hongqi and Sun, Xian},
  journal={IEEE Transactions on Geoscience and Remote Sensing},
  volume={61},
  pages={1--12},
  year={2023},
  publisher={IEEE}
}

@inproceedings{zhou2024migc,
  title={Migc: Multi-instance generation controller for text-to-image synthesis},
  author={Zhou, Dewei and Li, You and Ma, Fan and Zhang, Xiaoting and Yang, Yi},
  booktitle={Proceedings of the IEEE/CVF conference on computer vision and pattern recognition},
  pages={6818--6828},
  year={2024}
}

@inproceedings{zheng2023layoutdiffusion,
  title={Layoutdiffusion: Controllable diffusion model for layout-to-image generation},
  author={Zheng, Guangcong and Zhou, Xianpan and Li, Xuewei and Qi, Zhongang and Shan, Ying and Li, Xi},
  booktitle={Proceedings of the IEEE/CVF Conference on Computer Vision and Pattern Recognition},
  pages={22490--22499},
  year={2023}
}

@inproceedings{wang2024instancediffusion,
  title={Instancediffusion: Instance-level control for image generation},
  author={Wang, Xudong and Darrell, Trevor and Rambhatla, Sai Saketh and Girdhar, Rohit and Misra, Ishan},
  booktitle={Proceedings of the IEEE/CVF conference on computer vision and pattern recognition},
  pages={6232--6242},
  year={2024}
}

@article{hou2025instada,
  title={InstaDA: Augmenting Instance Segmentation Data with Dual-Agent System},
  author={Hou, Xianbao and He, Yonghao and Boukhers, Zeyd and See, John and Su, Hu and Sui, Wei and Yang, Cong},
  journal={arXiv preprint arXiv:2509.02973},
  year={2025}
}

@inproceedings{fan2024divergen,
  title={Divergen: Improving instance segmentation by learning wider data distribution with more diverse generative data},
  author={Fan, Chengxiang and Zhu, Muzhi and Chen, Hao and Liu, Yang and Wu, Weijia and Zhang, Huaqi and Shen, Chunhua},
  booktitle={Proceedings of the IEEE/CVF Conference on Computer Vision and Pattern Recognition},
  pages={3986--3995},
  year={2024}
}

@inproceedings{xie2015holistically,
  title={Holistically-nested edge detection},
  author={Xie, Saining and Tu, Zhuowen},
  booktitle={Proceedings of the IEEE international conference on computer vision},
  pages={1395--1403},
  year={2015}
}

@inproceedings{
podell2024sdxl,
title={{SDXL}: Improving Latent Diffusion Models for High-Resolution Image Synthesis},
author={Dustin Podell and Zion English and Kyle Lacey and Andreas Blattmann and Tim Dockhorn and Jonas M{\"u}ller and Joe Penna and Robin Rombach},
booktitle={The Twelfth International Conference on Learning Representations},
year={2024},
url={https://openreview.net/forum?id=di52zR8xgf}
}

@inproceedings{
hu2022lora,
title={Lo{RA}: Low-Rank Adaptation of Large Language Models},
author={Edward J Hu and yelong shen and Phillip Wallis and Zeyuan Allen-Zhu and Yuanzhi Li and Shean Wang and Lu Wang and Weizhu Chen},
booktitle={International Conference on Learning Representations},
year={2022},
url={https://openreview.net/forum?id=nZeVKeeFYf9}
}

@inproceedings{zhang2023adding,
  title={Adding conditional control to text-to-image diffusion models},
  author={Zhang, Lvmin and Rao, Anyi and Agrawala, Maneesh},
  booktitle={Proceedings of the IEEE/CVF international conference on computer vision},
  pages={3836--3847},
  year={2023}
}

@inproceedings{zavadski2024controlnet,
  title={Controlnet-xs: Rethinking the control of text-to-image diffusion models as feedback-control systems},
  author={Zavadski, Denis and Feiden, Johann-Friedrich and Rother, Carsten},
  booktitle={European Conference on Computer Vision},
  pages={343--362},
  year={2024},
  organization={Springer}
}

@inproceedings{wang2025ficgen,
  title={FICGen: Frequency-Inspired Contextual Disentanglement for Layout-driven Degraded Image Generation},
  author={Wang, Wenzhuang and Zhao, Yifan and Ma, Mingcan and Liu, Ming and Jiang, Zhonglin and Chen, Yong and Li, Jia},
  booktitle={Proceedings of the IEEE/CVF International Conference on Computer Vision},
  pages={19097--19107},
  year={2025}
}

@article{diorrsvg,
  author={Zhan, Yang and Xiong, Zhitong and Yuan, Yuan},
  journal={IEEE Transactions on Geoscience and Remote Sensing}, 
  title={RSVG: Exploring Data and Models for Visual Grounding on Remote Sensing Data}, 
  year={2023},
  volume={61},
  number={},
  pages={1-13},
  doi={10.1109/TGRS.2023.3250471}
  }

@inproceedings{xia2018dota,
  title={DOTA: A large-scale dataset for object detection in aerial images},
  author={Xia, Gui-Song and Bai, Xiang and Ding, Jian and Zhu, Zhen and Belongie, Serge and Luo, Jiebo and Datcu, Mihai and Pelillo, Marcello and Zhang, Liangpei},
  booktitle={Proceedings of the IEEE conference on computer vision and pattern recognition},
  pages={3974--3983},
  year={2018}
}

@inproceedings{xie2023boxdiff,
  title={Boxdiff: Text-to-image synthesis with training-free box-constrained diffusion},
  author={Xie, Jinheng and Li, Yuexiang and Huang, Yawen and Liu, Haozhe and Zhang, Wentian and Zheng, Yefeng and Shou, Mike Zheng},
  booktitle={Proceedings of the IEEE/CVF International Conference on Computer Vision},
  pages={7452--7461},
  year={2023}
}

@inproceedings{dahary2024yourself,
  title={Be yourself: Bounded attention for multi-subject text-to-image generation},
  author={Dahary, Omer and Patashnik, Or and Aberman, Kfir and Cohen-Or, Daniel},
  booktitle={European Conference on Computer Vision},
  pages={432--448},
  year={2024},
  organization={Springer}
}

@inproceedings{rombach2022high,
  title={High-resolution image synthesis with latent diffusion models},
  author={Rombach, Robin and Blattmann, Andreas and Lorenz, Dominik and Esser, Patrick and Ommer, Bj{\"o}rn},
  booktitle={Proceedings of the IEEE/CVF conference on computer vision and pattern recognition},
  pages={10684--10695},
  year={2022}
}

@inproceedings{islam2024diffusemix,
  title={Diffusemix: Label-preserving data augmentation with diffusion models},
  author={Islam, Khawar and Zaheer, Muhammad Zaigham and Mahmood, Arif and Nandakumar, Karthik},
  booktitle={Proceedings of the IEEE/CVF Conference on Computer Vision and Pattern Recognition},
  pages={27621--27630},
  year={2024}
}

@inproceedings{zhao2023x,
  title={X-paste: Revisiting scalable copy-paste for instance segmentation using clip and stablediffusion},
  author={Zhao, Hanqing and Sheng, Dianmo and Bao, Jianmin and Chen, Dongdong and Chen, Dong and Wen, Fang and Yuan, Lu and Liu, Ce and Zhou, Wenbo and Chu, Qi and others},
  booktitle={International Conference on Machine Learning},
  pages={42098--42109},
  year={2023},
  organization={PMLR}
}

@inproceedings{li2023gligen,
  title={Gligen: Open-set grounded text-to-image generation},
  author={Li, Yuheng and Liu, Haotian and Wu, Qingyang and Mu, Fangzhou and Yang, Jianwei and Gao, Jianfeng and Li, Chunyuan and Lee, Yong Jae},
  booktitle={Proceedings of the IEEE/CVF conference on computer vision and pattern recognition},
  pages={22511--22521},
  year={2023}
}

@article{ho2020denoising,
  title={Denoising diffusion probabilistic models},
  author={Ho, Jonathan and Jain, Ajay and Abbeel, Pieter},
  journal={Advances in neural information processing systems},
  volume={33},
  pages={6840--6851},
  year={2020}
}

@article{kingma2013auto,
  title={Auto-encoding variational bayes},
  author={Kingma, Diederik P and Welling, Max},
  journal={arXiv preprint arXiv:1312.6114},
  year={2013}
}

@inproceedings{ronneberger2015u,
  title={U-net: Convolutional networks for biomedical image segmentation},
  author={Ronneberger, Olaf and Fischer, Philipp and Brox, Thomas},
  booktitle={International Conference on Medical image computing and computer-assisted intervention},
  pages={234--241},
  year={2015},
  organization={Springer}
}

@inproceedings{radford2021learning,
  title={Learning transferable visual models from natural language supervision},
  author={Radford, Alec and Kim, Jong Wook and Hallacy, Chris and Ramesh, Aditya and Goh, Gabriel and Agarwal, Sandhini and Sastry, Girish and Askell, Amanda and Mishkin, Pamela and Clark, Jack and others},
  booktitle={International conference on machine learning},
  pages={8748--8763},
  year={2021},
  organization={PmLR}
}

@article{li2020object,
  title={Object detection in optical remote sensing images: A survey and a new benchmark},
  author={Li, Ke and Wan, Gang and Cheng, Gong and Meng, Liqiu and Han, Junwei},
  journal={ISPRS journal of photogrammetry and remote sensing},
  volume={159},
  pages={296--307},
  year={2020},
  publisher={Elsevier}
}

@article{heusel2017gans,
  title={Gans trained by a two time-scale update rule converge to a local nash equilibrium},
  author={Heusel, Martin and Ramsauer, Hubert and Unterthiner, Thomas and Nessler, Bernhard and Hochreiter, Sepp},
  journal={Advances in neural information processing systems},
  volume={30},
  year={2017}
}

@article{ren2016faster,
  title={Faster R-CNN: Towards real-time object detection with region proposal networks},
  author={Ren, Shaoqing and He, Kaiming and Girshick, Ross and Sun, Jian},
  journal={IEEE transactions on pattern analysis and machine intelligence},
  volume={39},
  number={6},
  pages={1137--1149},
  year={2016},
  publisher={IEEE}
}

@inproceedings{xie2021oriented,
  title={Oriented R-CNN for object detection},
  author={Xie, Xingxing and Cheng, Gong and Wang, Jiabao and Yao, Xiwen and Han, Junwei},
  booktitle={Proceedings of the IEEE/CVF international conference on computer vision},
  pages={3520--3529},
  year={2021}
}

@inproceedings{li2021image,
  title={Image synthesis from layout with locality-aware mask adaption},
  author={Li, Zejian and Wu, Jingyu and Koh, Immanuel and Tang, Yongchuan and Sun, Lingyun},
  booktitle={Proceedings of the IEEE/CVF International Conference on Computer Vision},
  pages={13819--13828},
  year={2021}
}

@misc{yolov8,
author = {Glenn Jocher and Ayush Chaurasia and Jing Qiu},
  title = {Ultralytics YOLO},
  version = {8.0.0},
  year = {2023},
  url = {https://github.com/ultralytics/ultralytics},
  license = {AGPL-3.0}
}

@inproceedings{jia2024ssmg,
  title={Ssmg: Spatial-semantic map guided diffusion model for free-form layout-to-image generation},
  author={Jia, Chengyou and Luo, Minnan and Dang, Zhuohang and Dai, Guang and Chang, Xiaojun and Wang, Mengmeng and Wang, Jingdong},
  booktitle={Proceedings of the AAAI Conference on Artificial Intelligence},
  volume={38},
  number={3},
  pages={2480--2488},
  year={2024}
}


\section*{Supplementary material}

# RSGen: Enhancing Layout-Driven Remote Sensing Image Generation with Diverse Edge Guidance

## Supplementary Material

## 1 Implementation Details

The section provides additional implementation details. We first describe the basic experimental configuration. Then we present the training settings for LoRA [7] and the YOLOv8 [8] model. After that, we introduce the evaluation for the Fréchet Inception Distance (FID) [6] and the cross-validation details used in our experiments. Finally, we provide the detailed training configurations for MMDetection [1] and MMRotate [28].

### 1.1 Experimental Configuration

We implement the training for the baseline models (CC-Diff [26], MIGC [27], and FICGen [19]) as well as our FGControl module using PyTorch 2.1.0 and CUDA 12.1.0 on 8 NVIDIA H20 GPUs. The training follows the official settings of the respective methods. To ensure a fair comparison, we train FICGen for 100 epochs in our main experiments. We provide the evaluation of FICGen under its official 300 epochs training setting in the Additional Experimental Results section. Furthermore, we evaluate all metrics, including FID and YOLOScore [10], using 8 NVIDIA L20 GPUs with PyTorch 2.5.1 and CUDA 12.4

### 1.2 LoRA Training Details

We fine-tune the SDXL [13] base model using the standard Low-Rank Adaptation (LoRA). The training dataset consists of the composited edge maps detailed in the main text. We freeze the text encoders and exclusively train the U-Net parameters [16]. We set the LoRA rank (dimension) to 64 and the network alpha to 8. The input training images are processed at a resolution of $1024 \times 1024$. For the optimization, we utilize the AdamW [12] optimizer with a learning rate of 0.0004, managed by a cosine learning rate scheduler with a warmup phase. The training is conducted for a total of 7,500 steps with a batch size of 4 and 4 gradient accumulation steps. Furthermore, we employ FP16 mixed precision to accelerate the process and apply Min-SNR [4] weighting with a gamma value of 5 to improve convergence stability.

### 1.3  YOLOScore Training Details

We train distinct YOLO detectors tailored to the bounding box formats of each dataset. For the DIOR-RSVG [24] dataset, which relies on horizontal bounding boxes (HBB), we employ the YOLOv8 Nano (yolov8n). Conversely, for the DOTA-v1.0 [20] and the HRSC2016 [11] datasets (detailed in the Additional Experimental Results section), which require oriented bounding boxes (OBB), we utilize the YOLOv8 Medium OBB (yolov8m-obb).

Across all datasets, models are trained on their respective train splits for 50 epochs using a batch size of 16 and an input image resolution of $512 \times 512$. To evaluate layout consistency, images generated from specific spatial layouts are assessed against the corresponding ground-truth annotations. Because test set annotations are unavailable for DIOR-RSVG and DOTA-v1.0, we utilize their val sets for both generation and evaluation. For HRSC2016, we use the test set for both generation and evaluation. During these validation phases, we adjust the parameters to align with the settings of CC-Diff: the DIOR-RSVG evaluation image size is scaled to $800 \times 800$, while DOTA-v1.0 and HRSC2016 are maintained at $512 \times 512$, all with a batch size of 16.

### 1.4  FID Computation

The FID is computed using a pre-trained Inception-v3 network [17]. The real images from the DIOR-RSVG dataset are resized to $800 \times 800$, while images from both the DOTA-v1.0 and HRSC2016 datasets are resized to $512 \times 512$.

### 1.5  Cross-Validation

To evaluate generalization, models trained on DIOR-RSVG are tested on DOTA-v1.0. We align the differing category sets by mapping shared DOTA categories to their DIOR-RSVG equivalents (e.g., merging "small-vehicle" and "large-vehicle" into "vehicle"). Annotations for unshared categories are filtered out to ensure a standardized and fair evaluation.

### 1.6  MMDetection and MMRotate Training Details

To evaluate downstream object detection performance, all models are implemented using mmcv-full 1.7.2, MMDetection 2.28.2, and MMRotate 0.3.4. For HBB tasks on the DIOR-RSVG and DOTA datasets, we utilize Faster R-CNN [14], resizing input images to $800 \times 800$ and applying standard data augmentations such as random horizontal flipping with a 0.5 probability. For OBB detection, architectures and angle representations are tailored to the specific datasets. Specifically, for DOTA-v1.0, we train a Rotated Faster R-CNN [21] with a ResNet-50 backbone [5] using the le90 angle definition and a $1024 \times 1024$ image resolution. Conversely, for the HRSC2016 dataset, we employ S2ANet [3] with a ResNet-50 backbone, utilizing the le135 angle definition and an $800 \times 800$ resolution. Both OBB configurations incorporate comprehensive spatial augmentations, including random horizontal, vertical, and diagonal flipping.

**Table 1:** Comparison of training times. Both FGControl and LoRA fine-tuning require significantly less time than the base L2I model.

| Component | Time |
|---|---|
| Base Model | 14h 43m |
| FGControl (Ours) | 8h 31m |
| LoRA (Ours) | 4h 40m |

**Table 2:** Inference efficiency on the HRSC dataset. While the Edge2Edge module incurs computational costs, adding FGControl introduces negligible overhead.

| Model Config. | VRAM (GB) | Latency (s) |
|---|---|---|
| Base Model | 10.18 | 9.58 |
| + FGControl | 10.19 | 10.16 |
| Edge2Edge | 46.05 | 41.04 |

## 2   Efficiency Analysis

In this section, we provide a detailed efficiency analysis of the RSGen framework. First, we report the training time required for the FGControl module and fine-tuning the base model via LoRA. Subsequently, we evaluate the inference efficiency in terms of latency and peak memory utilization upon integrating our proposed modules.

### 2.1   Training Time

To demonstrate the training efficiency of the proposed RSGen framework, we compare the training overhead of our modules against the base layout-to-image (L2I) model. As detailed in  Tab. 1, training the base model requires approximately 14 hours and 43 minutes. In contrast, optimizing the lightweight FGControl module takes only 8 hours and 31 minutes. Furthermore, fine-tuning the SDXL model via LoRA for the Edge2Edge module is highly efficient, requiring approximately 4 hours. These results indicate that our plug-and-play framework significantly enhances fine-grained layout control without introducing a prohibitive training burden.

### 2.2   Inference Efficiency

As shown in  Tab. 2, evaluations on the HRSC dataset reveal that integrating the lightweight FGControl module adds virtually no burden: VRAM consumption increases by a mere 0.01 GB, and latency rises by only 0.58 seconds per image. Furthermore, while the Edge2Edge module (operating at 50 steps with a 0.6 denoising strength) requires 46.05 GB of VRAM and approximately 41 seconds per image, it serves strictly as a one-time offline data augmentation step, introducing zero overhead during downstream inference.

## 3   Additional Experimental Results

This section provides further experimental evidence to validate the proposed RSGen framework, including its generalization on the HRSC2016 dataset, effectiveness on performance-saturated models, additional ablation studies on core modules, and justification for the selection of SDXL.

**Table 3:** Performance comparison on the HRSC2016 dataset. Equipped with RSGen, the base model achieves superior layout consistency and downstream detection performance in complex backgrounds.

| Method | FID ↓ | YOLOScore (OBB) ↑ | | mAP ↑ |
|---|---|---|---|---|
| | | $mAP_{50}$ | $mAP_{50-95}$ | |
| Ori (Real Data Only) | - | - | - | 37.71 |
| FICGen | 101.27 | 89.8 | 57.1 | 36.55 |
| FICGen + Ours | **99.85** | **90.8** | **61.2** | **38.39** |

**Table 4:** Evaluation on a performance-saturated model. Even when the base FICGen model is trained for 300 epochs to reach its performance bottleneck, integrating RSGen provides further improvements in layout consistency.

| Method | FID ↓ | YOLOScore (OBB) ↑ | |
|---|---|---|---|
| | | $mAP_{50}$ | $mAP_{50-95}$ |
| FICGen | **41.23** | **77.2** | 52.8 |
| FICGen + Ours | 41.41 | **77.2** | **54.8** |

### 3.1   Results on the HRSC2016 Dataset

To evaluate the generalization capability of the proposed RSGen framework, we conduct experiments on the HRSC2016 dataset. The dataset focuses on ship detection in satellite images characterized by highly complex backgrounds. Using FICGen as the base L2I model, we assess FID, YOLOScore, and mAP.

As shown in  Tab. 3, integrating RSGen simultaneously improves FID (from 101.27 to 99.85) and YOLOScore, specifically achieving a $+4.1$ gain in $mAP_{50-95}$. Crucially, evaluating the downstream detection performance reveals that simply augmenting the training set with FICGen degrades performance to 36.55 mAP, compared to 37.71 mAP when using solely real data. In contrast, incorporating RSGen significantly elevates generated data quality, reversing this degradation and boosting the overall mAP to 38.39. This confirms that our fine-grained control preserves structural integrity, enabling substantial gains for detectors even under severe background interference.

### 3.2   Improving Upon Performance-Saturated Model

To investigate the effectiveness of our framework when applied to a model with strong inherent control capabilities, we evaluate RSGen on a baseline that has reached performance saturation. Specifically, FICGen is utilized as the base L2I model and trained on the DOTA dataset. The baseline is trained for 300 epochs to ensure its performance reaches saturation. FID and YOLOScore are then assessed under the OBB setting. As shown in  Tab. 4, while the FID experiences a negligible increase (from 41.23 to 41.41), integrating RSGen yields a substantial

**Table 5:** Ablation study on different control modules. Compared to ControlNet and ControlNet-XS, our FGControl achieves superior layout consistency and fidelity.

| Method | FID $\downarrow$ | YOLOScore $\uparrow$ | |
| --- | --- | --- | --- |
| | | $mAP_{50}$ | $mAP_{50-95}$ |
| ControlNet | 128.04 | 50.5 | 37.5 |
| ControlNet-XS | 112.43 | 62.1 | 45.5 |
| FGControl (Ours) | **68.40** | **69.5** | **46.9** |

**Table 6:** Ablation on Scale Balanced Region Attention. Compared to standard SDXL I2I generation and Be Yourself, our method achieves the best layout consistency and generation fidelity.

| Method | FID $\downarrow$ | YOLOScore $\uparrow$ | |
| --- | --- | --- | --- |
| | | $mAP_{50}$ | $mAP_{50-95}$ |
| SDXL I2I | 69.18 | 67.3 | 44.7 |
| Be Yourself | 68.44 | 69.3 | 46.9 |
| Ours | **68.12** | **70.1** | **47.7** |

+2.0 gain in the $mAP_{50-95}$ metric, alongside a stable $mAP_{50}$. The improvement demonstrates that our fine-grained edge guidance effectively elevates the generation precision and overall performance ceiling, even for fully trained models.

### 3.3 Additional Ablation Studies

**Ablation on Control Modules.** To evaluate the proposed FGControl, we compare it against the established global control mechanisms ControlNet [25] and ControlNet-XS [23]. As detailed in Tab. 5, FGControl outperforms these global approaches across all metrics. Specifically, our module achieves a highly competitive FID of 68.40 and a YOLOScore ($mAP_{50-95}$) of 46.9. Our approach yields a substantial improvement in generation fidelity alongside consistent gains in spatial alignment accuracy. This confirms that confining high-frequency structural guidance within the layout bounding boxes achieves fine-grained local control without interfering with the global semantic generation.

**Ablation on Scale-Balanced Region Attention.** We evaluate the proposed Scale-Balanced Region Attention within the Edge2Edge module to assess its role in generating structural priors. The generation process utilizes 50 inference steps with a denoising strength of 0.6. To ensure a fair comparison, we adopt the hyperparameters established by Be Yourself [2]. Specifically, we utilize a dynamic step size that decays from 8 to 2, and we scale the maximum guidance iterations per step to 9 (proportionally adjusted from the original 15 to account for our 0.6 denoising strength).

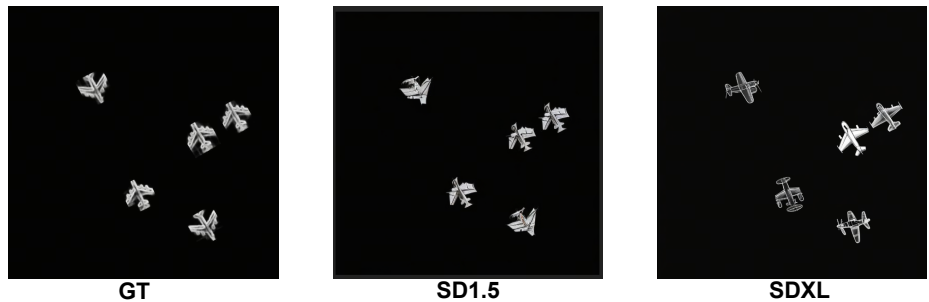

**Fig. 1:** Qualitative comparison of standard I2I generation (without LoRA) between SD 1.5 and SDXL. SDXL successfully preserves the structural integrity while introducing meaningful diversity.

As detailed in Tab. 6, standard SDXL I2I generation yields a low layout consistency (mAP$_{50-95}$ of 44.7) due to a lack of explicit spatial constraints, leading to semantic misalignment and boundary overflow. While Be Yourself introduces spatial guidance and improves the mAP$_{50-95}$ to 46.9, it suffers from an optimization bias that favors larger bounding boxes. Our area constraints mitigate this bias during latent updates, boosting mAP$_{50-95}$ to 47.7 while securing the best FID (68.12). These results demonstrate that our method ensures more balanced and precise spatial control across objects of varying scales, which provides the downstream FGControl module with accurate structural priors.

### 3.4 SDXL Selection

To justify the foundational model choice for the Edge2Edge module, we qualitatively compare the generation capabilities of SD 1.5 [15] and SDXL. For a fair evaluation, we conduct standard I2I generation using both models.

As illustrated in Fig. 1, SD 1.5 struggles to preserve the structural integrity of the original instances. It frequently fails to maintain the basic shape of the airplane and introduces noticeable noise. In contrast, SDXL successfully preserves the accurate structural contours of the airplane while concurrently introducing meaningful structural diversity. The qualitative comparison demonstrates that SDXL possesses a vastly superior inherent understanding of complex structures, making it the most suitable and robust foundation for generating high-quality edge priors in our framework.

## 4 Limitations and Long-tail Cases

Despite its robust performance, our method has certain limitations. As an auxiliary plug-and-play module, RSGen maximizes the layout consistency of existing L2I models but cannot fundamentally alter the inherent generative limits of the base model. Fig. 2 illustrates specific instances across different base models.

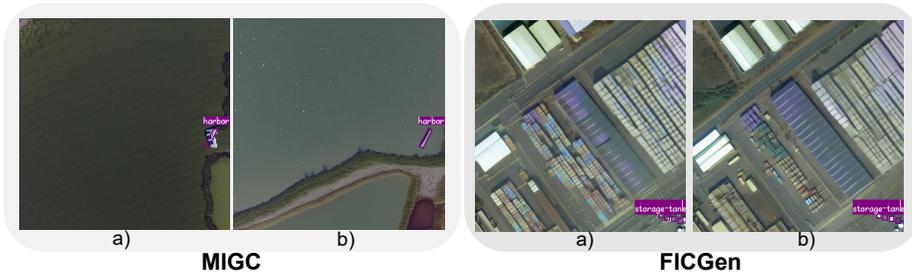

**Fig. 2:** Qualitative comparison of standard I2I generation (without LoRA) between SD 1.5 and SDXL. SDXL successfully preserves the structural integrity while introducing meaningful diversity.

**Table 7:** Evaluation of RSGen on the GLIGEN baseline. The modified U-Net architecture of GLIGEN interferes with our feature injection mechanism, leading to suboptimal performance.

| Method | FID ↓ | YOLOScore ↑ | |
| --- | --- | --- | --- |
| | | mAP$_{50}$ | mAP$_{50-95}$ |
| GLIGEN | **77.08** | **78.6** | **50.8** |
| GLIGEN + Ours | 81.24 | 49.8 | 32.2 |

The MIGC results (Left) highlight the limitations of the base model in maintaining context coherence. Specifically, the generated harbor in image (a) fails to blend semantically with the surrounding environment. Adding RSGen (image b) successfully constrains the harbor within the box but fails to resolve it. The FICGen results (Right) demonstrate poor performance when handling extremely small objects (image a). The combination of RSGen in image (b) somewhat improves object generation but cannot completely resolve minute visual details.

Furthermore, as detailed in Tab. 7, we observe suboptimal performance when integrating RSGen with GLIGEN [9]. This is primarily attributed to the modified U-Net architecture of GLIGEN, which inserts an additional Gated Self-Attention layer between the standard attention modules, thereby interfering with our feature injection mechanism. Finally, while RSGen is designed as a plug-and-play module within the *diffusers* library, it cannot be directly integrated into specialized remote sensing models, such as AeroGen [18] and OF-Diff [22].

## 5  Qualitative Analysis

To further demonstrate the effectiveness of our proposed framework, we provide qualitative visualizations from two perspectives: the quality of the generated images and the resulting improvements in downstream detection tasks.

First, we showcase the high-quality synthetic data generated by RSGen. As illustrated in Fig. 3, we present a three-column visualization consisting of the in-

put spatial layouts, the diverse edge maps generated by our Edge2Edge module, and the final images generated by the L2I model. These results demonstrate that our framework produces highly realistic remote sensing data that strictly adheres to the given spatial constraints with pixel-level precision, successfully capturing both global layouts and fine-grained structural details. Notably ( Fig. 3, fifth row), the decoupling mechanism in FGControl naturally reduces reliance on guidance when edge maps are blurred. This prevents instance generation failure and ensures robust spatial alignment despite low-quality edges.

Second, we present visual comparisons of detection performance to highlight the practical benefits of our framework, with FICGen serving as the base model. As shown in  Fig. 4, we perform HBB detection on the DIOR-RSVG dataset and OBB detection on the DOTA dataset. The visualization confirms that augmenting the training set with our generated data significantly enhances bounding box precision and overall recognition capability.

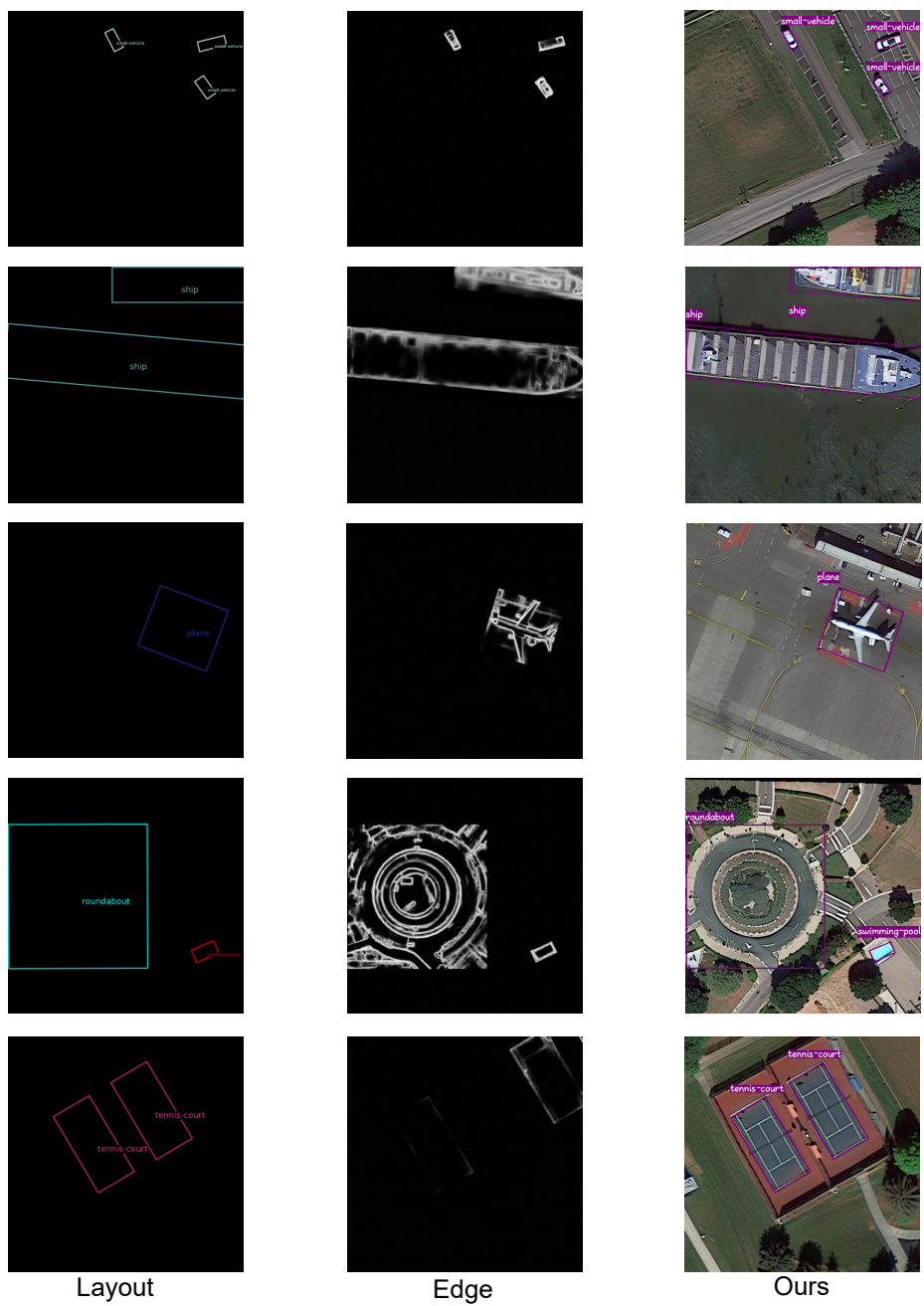

Layout                        Edge                         Ours

**Fig. 3:** Visualization of the RSGen image generation process. Images generated by RSGen demonstrate the ability of the framework to maintain strict spatial alignment and high visual fidelity.

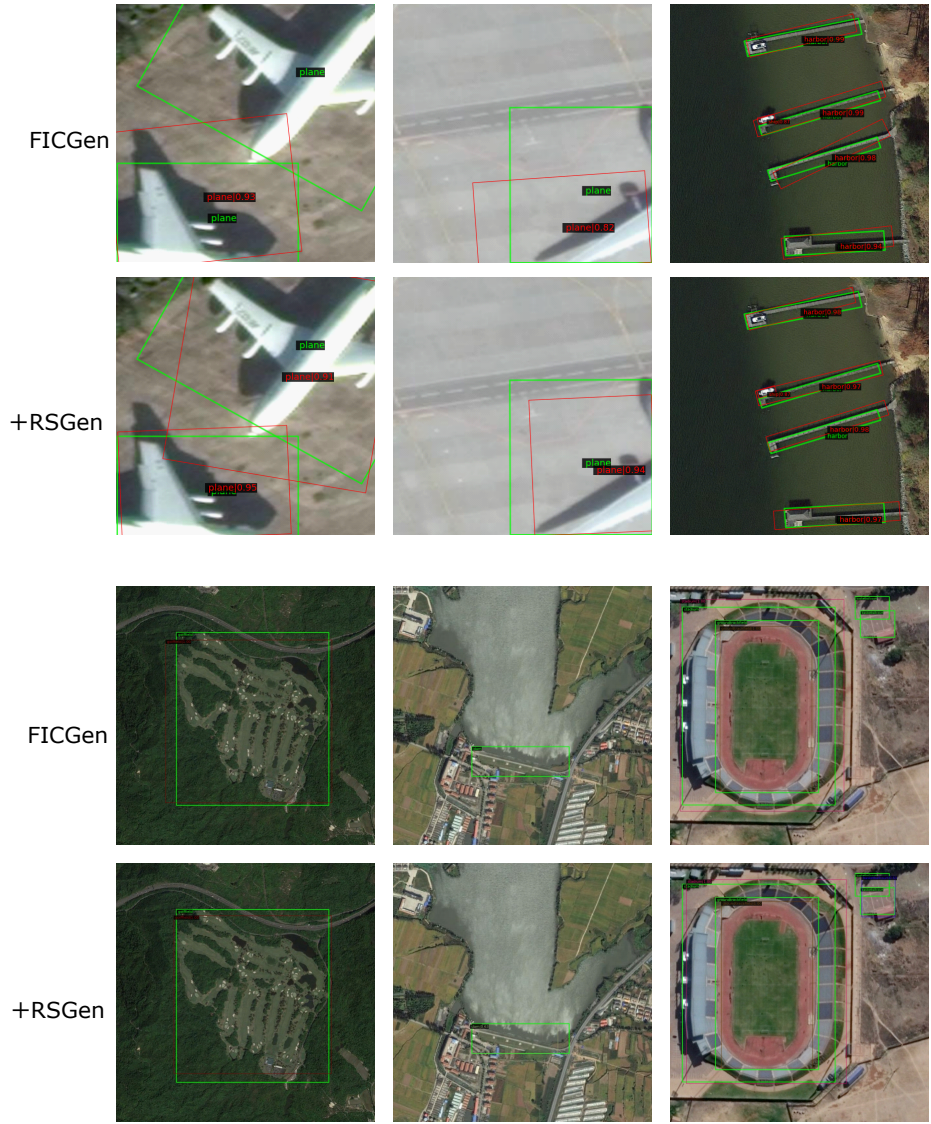

**Fig. 4:** Visual comparison of detection results on DIOR-RSVG (HBB) and DOTA (OBB) datasets using FICGen. Models trained with our augmented data exhibit higher localization precision and improved recognition of objects.

# References

1. Chen, K., Wang, J., Pang, J., Cao, Y., Xiong, Y., Li, X., Sun, S., Feng, W., Liu, Z., Xu, J., Zhang, Z., Cheng, D., Zhu, C., Cheng, T., Zhao, Q., Li, B., Lu, X., Zhu, R., Wu, Y., Dai, J., Wang, J., Shi, J., Ouyang, W., Loy, C.C., Lin, D.: MMDetection: Open mmlab detection toolbox and benchmark. arXiv preprint arXiv:1906.07155 (2019)

2. Dahary, O., Patashnik, O., Aberman, K., Cohen-Or, D.: Be yourself: Bounded attention for multi-subject text-to-image generation. In: European Conference on Computer Vision. pp. 432–448. Springer (2024)

3. Han, J., Ding, J., Li, J., Xia, G.S.: Align deep features for oriented object detection. IEEE transactions on geoscience and remote sensing **60**, 1–11 (2021)

4. Hang, T., Gu, S., Li, C., Bao, J., Chen, D., Hu, H., Geng, X., Guo, B.: Efficient diffusion training via min-snr weighting strategy. In: Proceedings of the IEEE/CVF International Conference on Computer Vision (ICCV). pp. 7441–7451 (October 2023)

5. He, K., Zhang, X., Ren, S., Sun, J.: Deep residual learning for image recognition. In: Proceedings of the IEEE conference on computer vision and pattern recognition. pp. 770–778 (2016)

6. Heusel, M., Ramsauer, H., Unterthiner, T., Nessler, B., Hochreiter, S.: Gans trained by a two time-scale update rule converge to a local nash equilibrium. Advances in neural information processing systems **30** (2017)

7. Hu, E.J., yelong shen, Wallis, P., Allen-Zhu, Z., Li, Y., Wang, S., Wang, L., Chen, W.: LoRA: Low-rank adaptation of large language models. In: International Conference on Learning Representations (2022), `https://openreview.net/forum?id=nZeVKeeFYf9`

8. Jocher, G., Chaurasia, A., Qiu, J.: Ultralytics yolo (2023), `https://github.com/ultralytics/ultralytics`

9. Li, Y., Liu, H., Wu, Q., Mu, F., Yang, J., Gao, J., Li, C., Lee, Y.J.: Gligen: Open-set grounded text-to-image generation. In: Proceedings of the IEEE/CVF conference on computer vision and pattern recognition. pp. 22511–22521 (2023)

10. Li, Z., Wu, J., Koh, I., Tang, Y., Sun, L.: Image synthesis from layout with locality-aware mask adaption. In: Proceedings of the IEEE/CVF International Conference on Computer Vision. pp. 13819–13828 (2021)

11. Liu, Z., Wang, H., Weng, L., Yang, Y.: Ship rotated bounding box space for ship extraction from high-resolution optical satellite images with complex backgrounds. IEEE geoscience and remote sensing letters **13**(8), 1074–1078 (2016)

12. Loshchilov, I., Hutter, F.: Decoupled weight decay regularization. arXiv preprint arXiv:1711.05101 (2017)

13. Podell, D., English, Z., Lacey, K., Blattmann, A., Dockhorn, T., Müller, J., Penna, J., Rombach, R.: SDXL: Improving latent diffusion models for high-resolution image synthesis. In: The Twelfth International Conference on Learning Representations (2024), `https://openreview.net/forum?id=di52zR8xgf`

14. Ren, S., He, K., Girshick, R., Sun, J.: Faster r-cnn: Towards real-time object detection with region proposal networks. IEEE transactions on pattern analysis and machine intelligence **39**(6), 1137–1149 (2016)

15. Rombach, R., Blattmann, A., Lorenz, D., Esser, P., Ommer, B.: High-resolution image synthesis with latent diffusion models. In: Proceedings of the IEEE/CVF conference on computer vision and pattern recognition. pp. 10684–10695 (2022)

16. Ronneberger, O., Fischer, P., Brox, T.: U-net: Convolutional networks for biomedical image segmentation. In: International Conference on Medical image computing and computer-assisted intervention. pp. 234–241. Springer (2015)

17. Szegedy, C., Vanhoucke, V., Ioffe, S., Shlens, J., Wojna, Z.: Rethinking the inception architecture for computer vision. In: Proceedings of the IEEE conference on computer vision and pattern recognition. pp. 2818–2826 (2016)

18. Tang, D., Cao, X., Wu, X., Li, J., Yao, J., Bai, X., Jiang, D., Li, Y., Meng, D.: Aerogen: Enhancing remote sensing object detection with diffusion-driven data generation. In: Proceedings of the Computer Vision and Pattern Recognition Conference. pp. 3614–3624 (2025)

19. Wang, W., Zhao, Y., Ma, M., Liu, M., Jiang, Z., Chen, Y., Li, J.: Ficgen: Frequency-inspired contextual disentanglement for layout-driven degraded image generation. In: Proceedings of the IEEE/CVF International Conference on Computer Vision. pp. 19097–19107 (2025)

20. Xia, G.S., Bai, X., Ding, J., Zhu, Z., Belongie, S., Luo, J., Datcu, M., Pelillo, M., Zhang, L.: Dota: A large-scale dataset for object detection in aerial images. In: Proceedings of the IEEE conference on computer vision and pattern recognition. pp. 3974–3983 (2018)

21. Xie, X., Cheng, G., Wang, J., Yao, X., Han, J.: Oriented r-cnn for object detection. In: Proceedings of the IEEE/CVF international conference on computer vision. pp. 3520–3529 (2021)

22. Ye, Z., Ma, S., Yang, J., Yang, X., Gong, Z., Yang, X., Wang, H.: Object fidelity diffusion for remote sensing image generation. In: The Fourteenth International Conference on Learning Representations (2026), `https://openreview.net/forum?id=ngfIm9aPsH`

23. Zavadski, D., Feiden, J.F., Rother, C.: Controlnet-xs: Rethinking the control of text-to-image diffusion models as feedback-control systems. In: European Conference on Computer Vision. pp. 343–362. Springer (2024)

24. Zhan, Y., Xiong, Z., Yuan, Y.: Rsvg: Exploring data and models for visual grounding on remote sensing data. IEEE Transactions on Geoscience and Remote Sensing **61**, 1–13 (2023). `https://doi.org/10.1109/TGRS.2023.3250471`

25. Zhang, L., Rao, A., Agrawala, M.: Adding conditional control to text-to-image diffusion models. In: Proceedings of the IEEE/CVF international conference on computer vision. pp. 3836–3847 (2023)

26. Zhang, M., Liu, Y., Liu, Y., Zhao, Y., Ye, Q.: Cc-diff: enhancing contextual coherence in remote sensing image synthesis. arXiv preprint arXiv:2412.08464 (2024)

27. Zhou, D., Li, Y., Ma, F., Zhang, X., Yang, Y.: Migc: Multi-instance generation controller for text-to-image synthesis. In: Proceedings of the IEEE/CVF conference on computer vision and pattern recognition. pp. 6818–6828 (2024)

28. Zhou, Y., Yang, X., Zhang, G., Wang, J., Liu, Y., Hou, L., Jiang, X., Liu, X., Yan, J., Lyu, C., Zhang, W., Chen, K.: Mmrotate: A rotated object detection benchmark using pytorch. In: Proceedings of the 30th ACM International Conference on Multimedia (2022)


\end{document}